\documentclass[10pt,twocolumn,letterpaper]{article}

\usepackage{cvpr}              
\usepackage{graphicx}
\usepackage{amsmath,amssymb,amsfonts,dsfont,bm,bbm,mathrsfs,pifont}
\usepackage{xcolor}
\usepackage{colortbl}
\usepackage{algorithm}
\usepackage{algorithmicx}
\usepackage[noend]{algpseudocode}
\usepackage{booktabs}
\usepackage{array}

\newtheorem{Proposition}{Proposition}
\usepackage{bbding}
\makeatletter
\newcommand{\ssymbol}[1]{$^{\@fnsymbol{#1}}$}

\makeatother

\usepackage[colorlinks,linkcolor=blue]{hyperref}

\usepackage[capitalize]{cleveref}
\crefname{section}{Sec.}{Secs.}
\Crefname{section}{Section}{Sections}
\Crefname{table}{Table}{Tables}
\crefname{table}{T}{Tabs.}

\begin{document}
\renewcommand\arraystretch{1.5}
\title{Out-of-Distributed Semantic Pruning for Robust Semi-Supervised Learning } 

\newcommand*\samethanks[1][\value{footnote}]{\footnotemark[#1]}
\author{
    Yu Wang$^{1,4}$\thanks{Equal contribution.} \quad Pengchong Qiao$^{1,2,4}\samethanks$ \quad Chang Liu$^{3}$ \quad Guoli Song$^{2,4}$  \quad Xiawu Zheng$^{2,4}$\thanks{Corresponding author.} \quad Jie Chen$^{1,2,4}\samethanks$
\and
    $^{1}$ School of Electronic and Computer Engineering, Peking University, Shenzhen, China\\
$^{2}$ Peng Cheng Laboratory, Shenzhen, China\\
$^{3}$ Department of Automation and BNRist, Tsinghua University, Beijing, China \\
$^{4}$ AI for Science(AI4S)-Preferred Program,Peking University Shenzhen Graduate School, China \\
\tt\small \{rain\_wang,pcqiao\}@stu.pku.edu.cn, \{songgl,zhengxw01,chenj\}@pcl.ac.cn, \\ 
\tt\small \{liuchang2022\}@tsinghua.edu.cn
}
\maketitle

\begin{abstract}
Recent advances in robust semi-supervised learning (SSL) typically filter out-of-distribution (OOD) information at the sample level. We argue that an overlooked problem of robust SSL is its corrupted information on \textbf{semantic} level, practically limiting the development of the field. In this paper, we take an initial step to explore and propose a unified framework termed OOD Semantic Pruning (OSP), which aims at pruning OOD semantics out from in-distribution (ID) features. Specifically, (i) we propose an aliasing OOD matching module to pair each ID sample with an OOD sample with semantic overlap. (ii) We design a soft orthogonality regularization, which first transforms each ID feature by suppressing its semantic component that is collinear with paired OOD sample. It then forces the predictions before and after soft orthogonality decomposition to be consistent. Being practically simple, our method shows a strong performance in OOD detection and ID classification on challenging benchmarks. In particular, OSP surpasses the previous state-of-the-art by \textbf{13.7\%} on accuracy for ID classification and \textbf{5.9\%} on AUROC for OOD detection on TinyImageNet dataset. The source codes are publicly available at \href{https://github.com/rain305f/OSP}{https://github.com/rain305f/OSP}.

\end{abstract}

\begin{figure} 
\centering 
\includegraphics[width=\linewidth]{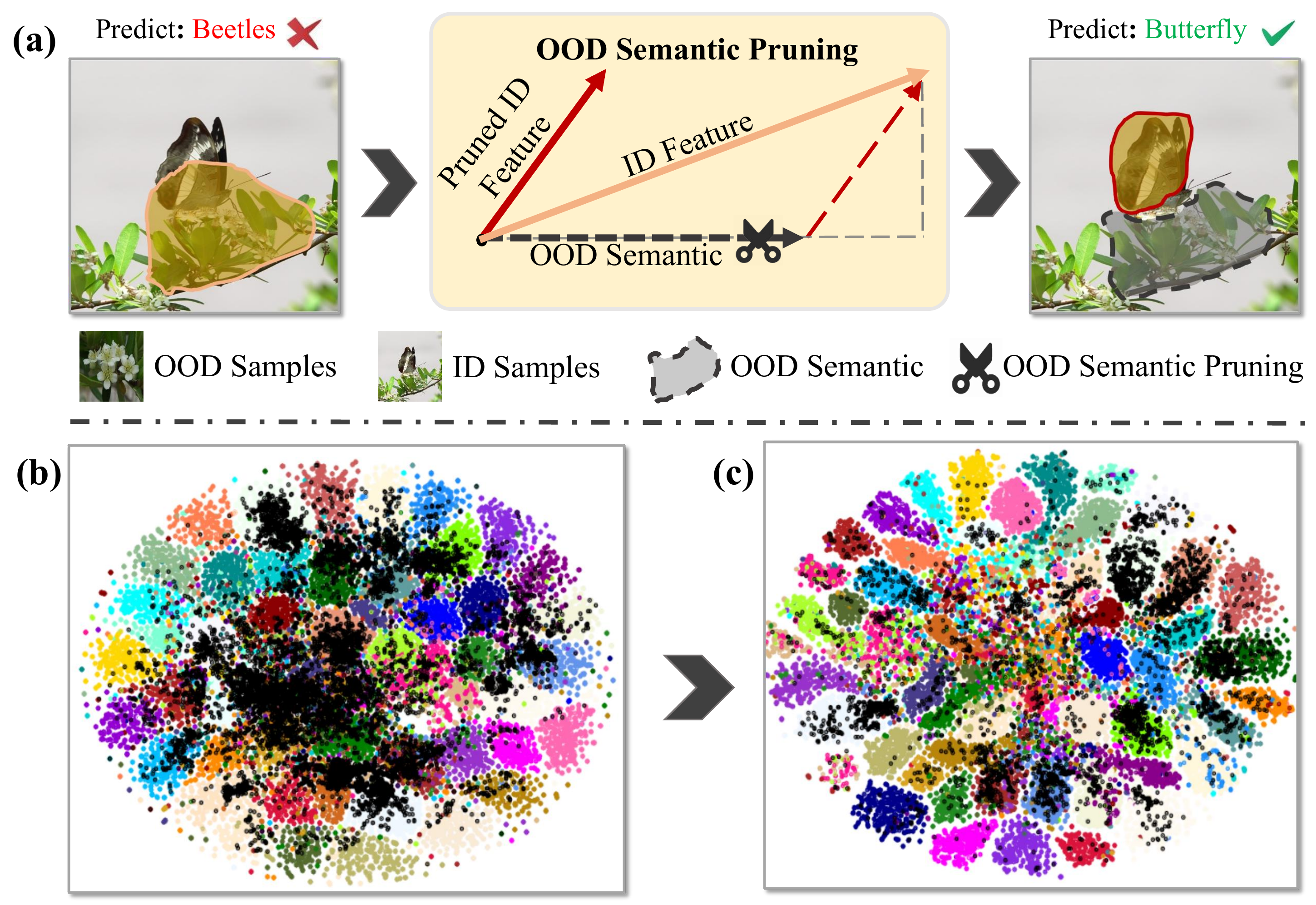} 
\caption{
(a) Intuitive diagram of OOD Semantic Pruning (OSP), that pruning OOD semantics out from ID features.
(b) t-SNE visualization~\cite{tsne} from the baseline~\cite{huang2021trash}.
(c) t-SNE visualization from our OSP model. The colorful dots donate ID features, while the black dots mark OOD features. The dots with the same color represent the features of the same class. Here, our OSP and the baseline are trained on CIFAR100 with 100 labeled data per class and 60\% OOD in unlabeled data.
} 
\label{Fig.motivation} 
\end{figure}

\section{Introduction}
\label{sec:intro}

Deep neural networks have obtained impressive performance on various tasks~\cite{krizhevsky2017imagenet,cheng2023m,cheng2023ssvmr,jin2022expectationmaximization,jin2023video,jin2023text,li2022toward,yao2023poserac,cheng2023parallel}. Their success is partially dependent on a large amount of labeled training data, of which the acquisition is expensive and time-consuming~\cite{AlecRadford2021LearningTV,iscen2019label,han2021unifying,cheng2023out,li2023acseg}. A prevailing way to reduce the dependency on human annotation is semi-supervised learning (SSL). It learns informative semantics using annotation-free and acquisition-easy unlabeled data to extend the label information from limited labeled data and has achieved promising results in various tasks~\cite{sohn2020fixmatch, xie2019unsupervised, AnttiTarvainen2017MeanTA, miyato2018virtual}. Unfortunately, classical SSL relies on a basic assumption that the labeled and unlabeled data are collected from the same distribution, which is difficult to hold in real-world applications.
In most practical cases, unlabeled data usually contains classes that are not seen in the labeled data. 
Existing works~\cite{chen2020semi,guo2020safe,oliver2018realistic,he2022safe} have shown that training the SSL model with these OOD samples in unlabeled data leads to a large degradation in performance.
To solve this problem, robust semi-supervised learning (Robust SSL) has been investigated to train a classification model that performs stably when the unlabeled set is corrupted by OOD samples. Typical methods focus on discarding OOD information at the sample level, that is, detecting and filtering OOD samples to purify the unlabeled set~\cite{chen2020semi,zhao2020robust,guo2020safe,he2022safe}. However, these methods ignore the semantic-level pollution caused by
the classification-useless semantics from OOD samples, which improperly disturbs the feature distribution learned from ID samples, eventually resulting in weak ID and OOD discrimination and low classification performance.
We provide an example to explain such a problem in Fig.~\ref{Fig.motivation}. 
As we can see, due to the semantics of \textit{Orchid} in OOD examples, the model pays too much attention to the background and misclassifies the \textit{Butterfly} as \textit{Beetle}.

In this paper, we propose Out-of-distributed Semantic Pruning (OSP) method to solve the problem mentioned above and achieve effective robust SSL.

Concretely, our OSP approach consists of two main modules. We first develop an aliasing OOD matching module to pair each ID sample with an OOD sample with which it has feature aliasing. Secondly, we propose a soft orthogonality regularization, which constrains the predictions of ID samples to keep consistent before and after soft-orthogonal decomposition according to their matching OOD samples.

We evaluate the effectiveness of our OSP in extensive robust semi-supervised image recognition benchmarks including MNIST~\cite{2011themnist}, CIFAR10~\cite{krizhevsky2009learning}, CIFAR100~\cite{krizhevsky2009learning} and TinyImageNet~\cite{JiaDeng2009ImageNetAL}. We show that our OSP obtains significant improvements compared to state-of-the-art alternatives (e.g., \textbf{13.7\%} and \textbf{15.0\%} on TinyImagetNet with an OOD ratio of 0.3 and 0.6 respectively).
Besides, we also empirically demonstrate that OSP indeed increases the feature discrimination between ID and OOD samples. To summarize, the contributions of this work are as follows:
\vspace{-0.2em}
\begin{itemize}

\item To the best of our knowledge, we are the first to exploit the OOD effects at the semantic level by regularization ID features to be orthogonal to OOD features.


\item We develop an aliasing OOD matching module that adaptively pairs each ID sample with an OOD sample.
In addition, we propose a soft orthogonality regularization to restrict ID and OOD features to be orthogonal.


\item We conduct extensive experiments on four datasets, i.e., MNIST, CIFAR10,  CIFAR100, and TIN200, and achieve new SOTA performance. Moreover, we analyze that the superiority of OSP lies in the enhanced discrimination between ID and OOD features.


\end{itemize}

\section{Related work}
\label{sec:related_work}
\subsection{Semi-Supervised Learning} 
Semi-supervised learning aims to learn informative semantics from unlabeled data to reduce the dependence on human annotations. Recently, many efforts have been made in SSL classification~\cite{YanbeiChen2022SemiSupervisedAU,DavidBerthelot2019MixMatchAH,DavidBerthelot2019ReMixMatchSL,KannalaJuho2019InterpolationCT,DongDongChen2018TrinetFS,SiyuanQiao2018DeepCF,ZhuoHuang2022UniversalSL,ZhaoweiCai2022SemisupervisedVT, TingChen2020BigSM,ZhongyiHan2020TowardsAA,YuFengLi2015TowardsMU,MehdiSajjadi2016RegularizationWS,EkinDCubuk2019AutoAugmentLA}. 
Powerful methods based on entropy minimization enforce their networks to make low-entropy predictions on unlabeled data~\cite{DongHyunLee2022PseudoLabelT,berthelot2019remixmatch,pham2021meta,9703190,DongHyunLee2022PseudoLabelT,YuFengLi2019SafeSL}. Another spectrum of popular approaches is consistency regularization, whose core idea is to obtain consistent prediction under various perturbations~\cite{miyato2018virtual,xie2019unsupervised, sohn2020fixmatch,AnttiTarvainen2017MeanTA,SamuliLaine2016TemporalEF}. VAT~\cite{miyato2018virtual} enforces prediction invariance under adversarial noises on unlabeled images.
UDA~\cite{xie2019unsupervised} and FixMatch~\cite{sohn2020fixmatch} employ weak and strong augmentation to compute the consistency loss.  

The effectiveness of these SSL methods relies on an assumption that the labeled and unlabeled data are drawn from the same distribution. 
However, in practice, such an assumption is difficult to satisfy, resulting in severe performance degeneration of close-set SSL~\cite{oliver2018realistic, zhao2020robust, LanZheGuo2022RobustDS}.
Thus, there is an urgent need to develop SSL algorithms that could work robustly with an unlabeled dataset that contains OOD samples. 

\begin{figure*} 
\centering 
\includegraphics[width=\textwidth]{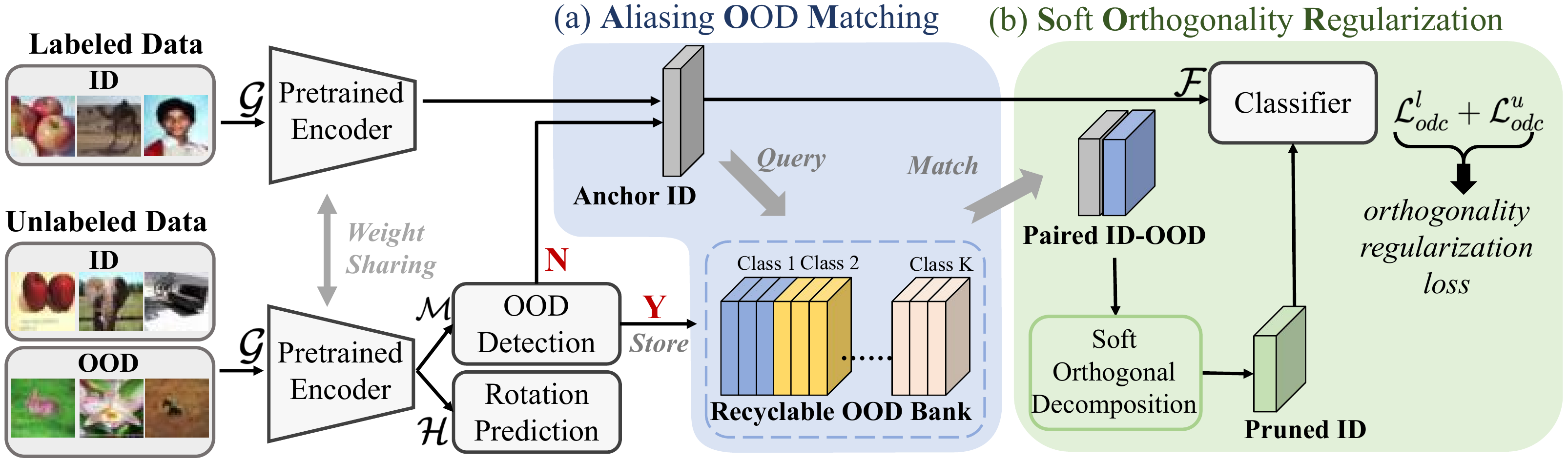} 
\caption{The overall architecture of our proposed OSP for robust semi-supervised classification. The core modules are aliasing OOD matching (AOM) and soft orthogonality regularization (SOR). The training process of our OSP consists of two-stage. At the pre-training stage, we pre-train the model with rotation prediction and K-ways predictor (Sec.~\ref{totalloss}). At the fine-tuning stage, we utilize the pre-trained OOD detection module to detect OOD samples in unlabeled data and store them in a class-wise memory bank, named recyclable OOD bank. To prune harmful OOD semantics out from ID features, the AOM selects an OOD sample with semantic overlap for each ID feature and composes ID-OOD pairs. Then the SOR applies a Soft Orthogonal Transform on ID-OOD pairs and generates pruned ID features. Finally, our proposed Orthogonality Regularization Loss constrains the predictions of ID features and corresponding pruned ID features to be consistent. During inference, the encoder and classifier are applied to K-ways ID classification. The details are shown in Sec.~\ref{sec:method}.} 

\label{Fig.method} 
\end{figure*}

\subsection{Robust Semi-Supervised Learning} 
Robust SSL aims to train a classification model that performs stably when the unlabeled set is corrupted by OOD samples~\cite{LanZheGuo2022RobustDS,MaximilianAugustin2020OutdistributionAS,KuniakiSaito2021OpenMatchOC,ZhuoHuang2022UniversalSL,KaidiCao2021OpenworldSL,DanHendrycks2016ABF,KiminLee2018ASU,JaehyungKim2020DistributionAR}.
This paper considers a common case: unlabeled data contains classes not seen in the labeled data~\cite{yu2020multi}.
Current typical approaches focus on removing the effects of OOD information at the sample-level~\cite{chen2020semi,zhao2020robust,guo2020safe,he2022safe}.
UASD~\cite{chen2020semi} utilizes self-distillation to detect OOD samples and filter them out later from unlabeled data.
MTC~\cite{yu2020multi} proposes a multi-task curriculum learning framework, which detects the OOD samples in unlabeled data and simultaneously estimates the probability of the sample belonging to OOD.
DS$^{3}$L~\cite{guo2020safe} trains a soft weighting function to assign small weights to OOD unlabeled samples and large weights to ID unlabeled samples.
More recently, some works have proposed utilizing OOD samples to improve the feature representation capacity of their models~\cite{luo2021empirical, huang2021trash}.
Simultaneously, they also inherited the idea of previous work to filter out OOD samples in classification supervision.
~\cite{luo2021empirical} extracts style features of ID samples and transfers OOD samples to ID style.
T2T~\cite{huang2021trash} employs an agent self-supervised task on both ID and OOD samples to enhance representation learning. Different from existing methods, we propose to prune the harmful OOD semantics out from ID features by regularizing ID and OOD features to be orthogonal, resulting in accurate ID classification and OOD detection.

\section{Method}
\label{sec:method}

\subsection{Preliminaries}

Give a small set of labeled data $\mathcal{D}^l=\{(x^l_i,y^l_i)\}^{N_l}_{i=1}$ and a large set of unlabeled data $\mathcal{D}^u = \{(x^u_i)\}^{N_u}_{i=1}$ ($N_l\ll N_u$), where $x^l_i,y^l_i$ and $x^u_i$ are the image and label of the
\textit{i}-th labeled data and the image of the \textit{i}-th unlabeled data.
The label space of labeled data contains $K$ labels, that is, $y^l_i \in C^l=\{1,...,K\}$.
The difference from classic SSL is that there exist OOD samples of unseen classes in the unlabeled training set.
Formally, $C^l \subset C^u $ and $C^{\text{OOD}} =C^u \setminus C^l $. Robust SSL aims to train a classification model that performs stably when the unlabeled set is corrupted by OOD samples. 

\subsection{Overview} 
    The architecture of our OSP is summarized in Fig.~\ref{Fig.method}.
    The previous state-of-the-art robust SSL method T2T \cite{huang2021trash} is selected as our baseline. 
    Following T2T, OSP has a shared encoder $\mathcal{G}(\cdot)$, a K-ways classifier $\mathcal{F}(\cdot)$, a rotation prediction head $\mathcal{H}(\cdot)$ and an OOD detection module $\mathcal{M}(\cdot)$. Different from T2T \cite{huang2021trash}, we design two novel modules, named aliasing OOD matching (AOM) and soft orthogonality regularization (SOR) respectively, to prune out-of-distributed semantic and obtain a robust classifier simultaneously. 
    The AOM module and SOR module are elaborated in Sec.~\ref{sec:ASMM} and Sec.~\ref{sec:SOR}, respectively.
    Inheriting the training paradigm of current robust SSL methods~\cite{yu2020multi, he2022safe, huang2021trash}, our OSP contains two training stages: the pre-training stage and fine-tuning stage, where the detailed descriptions are as follows.

\textbf{Pre-training stage.} The purpose of this stage is to obtain a pre-trained model that could detect OOD samples reasonably.
Following T2T \cite{huang2021trash}, we carry out a K-way classification on $\mathcal{D}^l$ and a self-supervised task \cite{MehdiNoroozi2016UnsupervisedLO} \cite{TingChen2020ASF} (i.e., rotation recognition\cite{SpyrosGidaris2018UnsupervisedRL}) on $\mathcal{D}^u$ to pre-train the encoder $\mathcal{G}(\cdot)$, the classifier $\mathcal{F}(\cdot)$, and the rotation predictor $\mathcal{H}(\cdot)$. 
Given a labeled input $x_i^l \in D^l$ and an unlabeled input $x_j^u \in  D^u $, we denote their representations as $z^l_i = \mathcal{G}(x^l_i)$ and $z_j^u = \mathcal{G}(x_j^u )$.
The training of model parameters is optimized by minimizing a supervised cross-entropy loss $\mathcal{L}_{ce}$ and a rotation loss $\mathcal{L}_{rot}$. Details are described in Sec.~\ref{totalloss}.

Meanwhile, we pre-train the OOD detection module $\mathcal{M}(\cdot)$ on $\mathcal{D}^l$ to calculate OOD scores $S(x^u)$ for unlabeled samples, which is used to distinguish ID samples and OOD samples in unlabeled data. Formally, we define the classifier as follows:
\begin{equation}
\label{ood}
g(x^u)=\left\{
\begin{aligned} &
\text{ID}, &\quad if~S(x^u) \ge \gamma , \\
&\text{OOD}, & \quad if~S(x^u) <  \gamma,
\end{aligned}
\right.
\end{equation}
where 
$\gamma$ is calculated by the Ostu algorithm\cite{NobuyukiOtsu1979ATS}  in our experiments\cite{huang2021trash}.
Additionally, we enforce our model to predict consistent predictions before and after adding Gaussian noises on feature maps $\mathcal{G}(\cdot)$, which helps to obtain more robust features.


\textbf{Fine-tuning stage.} 
The fine-tuning stage aims to refine the pre-trained model to obtain an accurate and robust classifier, which is achieved by the proposed AOM and SOR.

As illustrated in Fig.~\ref{Fig.method}, we first utilize the OOD detection module $\mathcal{M}(\cdot)$ to periodically split unlabeled data into subsets: ID unlabeled set and OOD unlabeled set, referring to \cite{huang2021trash}. The ID unlabeled set is then used to learn semantics from unlabeled data.
Due to OOD samples having conflicting targets with the classification, the compared baseline T2T~\cite{huang2021trash} drops the OOD unlabeled set. 
In contrast, we argue that the dropped set still contains useful information, which needs to be pruned in optimization. 
To this end, we propose the AOM and SOR to achieve such a purpose. 
Specifically, the AOM pairs each ID sample with an OOD sample with which it has feature aliasing. And then, the SOR constrains the predictions of ID samples to keep consistent before and after soft-orthogonal decomposition according to their matching OOD samples.

\subsection{Aliasing OOD Matching}
\label{sec:ASMM}
In this section, we introduce our aliasing OOD matching (AOM) Module and discuss how to select anchor ID samples and pair them with OOD samples with which they have feature aliasing.

\textbf{Anchor ID features.}
During training, we sample anchor ID images (queries) for each target category that appears in the current mini-batch. We denote the feature set of labeled candidate anchor images for category $c$ as $\mathcal{A}^l_c$, which contains features of labeled images with high confidence. Formally, 
\begin{equation}
\label{lquery}
\mathcal{A}^l_c = \{z^l_i | z^l_i = \mathcal{G}(x^l_i),y^l_i = c, p^l_i[c] > \delta \},
\end{equation}
where $y^l_i$, $z^l_i$ and $p^l_i$  are the ground-truth label, feature representation, and class probability for the labeled image $x^l_i$, respectively. 
Here, $\delta$ denotes the positive threshold and is set to 0.8 following \cite{huang2021trash}, and $p^l_i[c]$ is the predicted probability of class $c$.
For unlabeled data, counterpart $\mathcal{A}^u_c$ is computed as:
\begin{equation}
\label{uquery}
\mathcal{A}^u_c = \{z^u_i | z^u_i = \mathcal{G}(x^u_i),\hat y^u_i = c, max_{c}(p^u_i[c]) > \delta \},
\end{equation}
where $y^u_i = \mathop{\arg\max_{c}} (p^u_i[c])$ is the pseudo label of the image $x_{i}^{u}$.
This $\mathcal{A}^u_c$ is similar to $\mathcal{A}^l_c$, the only difference is that it uses the pseudo-label for class determination.
Based on $\mathcal{A}^l_c$ and $\mathcal{A}^u_c$, we obtain the set of all qualified ID anchors $\mathcal{A}_c$:
\begin{equation}
\label{allquery}
\mathcal{A}_c = \mathcal{A}^l_c \cup  \mathcal{A}^u_c.
\end{equation}

\begin{figure}[tbp] 
\centering 
\includegraphics[width=\linewidth]{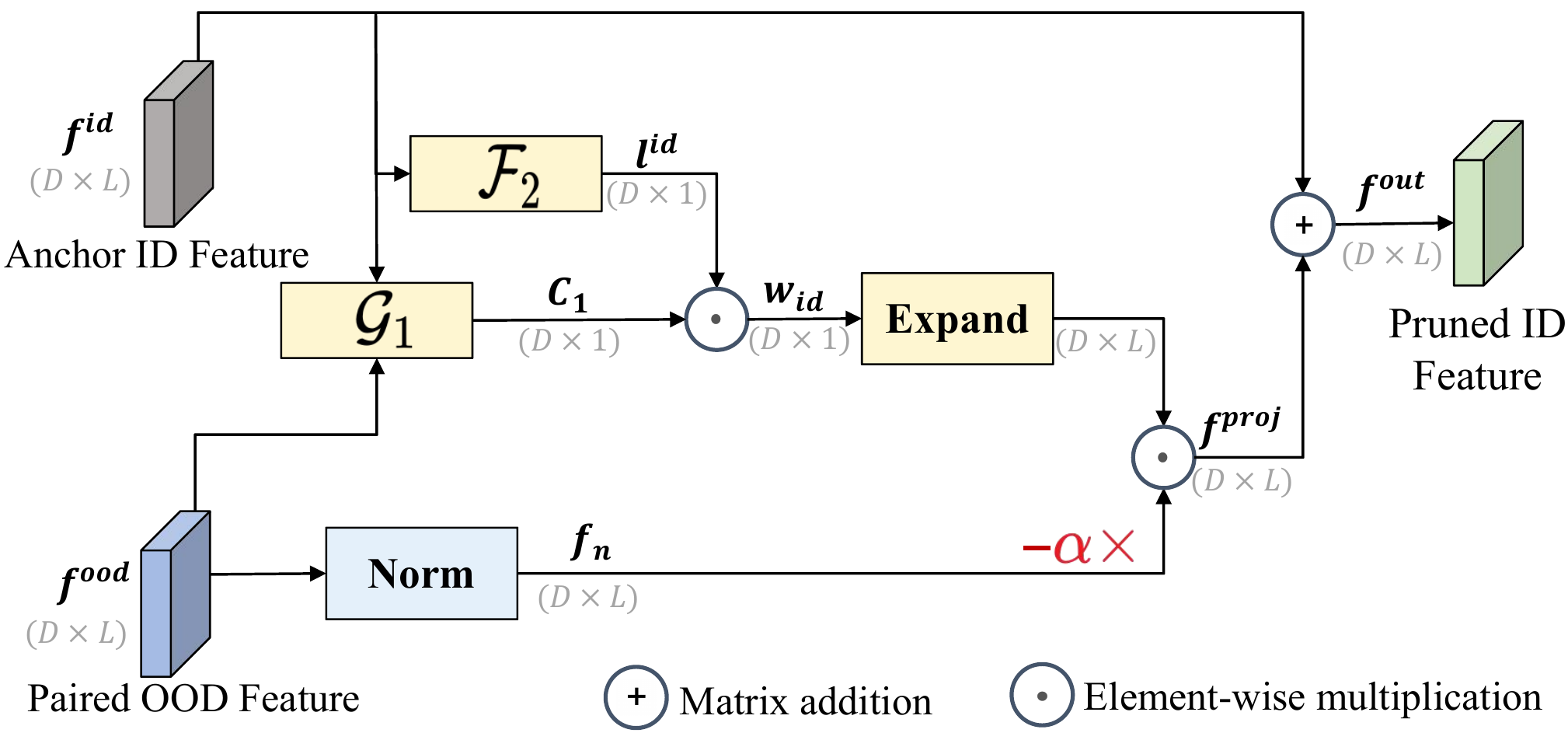} 
\caption{The pipeline of the soft orthogonal decomposition (SOD). The input of SOD is ID-OOD pairs, and its output is pruned ID features. Here, the function $\mathcal{G}_1$ calculates the cosine of the angle between two vectors, while the function $\mathcal{F}(\cdot)$ calculates the $l_2$-norm of a vector. First, we obtain the cosine of the angle of $f^{\text{ID}}$ and $f^{\text{OOD}}$, named as $C_1$ by $\mathcal{G}_1$. Then we get the $l_2$-norm $l^{\text{ID}}$ of anchor ID feature $f^{\text{ID}}$ by $\mathcal{F}(\cdot)$. We normalize anchor ID feature $f^{\text{ID}}$ and obtain $f_n$. Then we get the projector of $f^{\text{ID}}$ to $f^{\text{OOD}}$, named as $f^{\text{proj}}$. Finally, we get the pruned ID feature$f^{\text{out}} = f^{\text{ID}} - \alpha f^{\text{proj}}$. 
} 
\label{Fig.sor} 
\end{figure}

\textbf{Recyclable OOD samples.} We define a binary variable $n_i(c)$ to identify whether an unlabeled image $x^u_i \in \mathcal{D}^u$ is qualified to be a recyclable OOD sample of category $c$. For a target category $c$, a qualified recyclable OOD sample should highly probably belong to OOD samples and share class-agnostic features with ID samples belonging to the category $c$.
Therefore, $n_i(c)$ is formalized as follows:
\begin{equation}
\label{recycleood}
n_i(c) = \mathbbm{1}[\hat y^u_i= c]\cdot \mathbbm{1}[g(x^u_i) = \text{OOD}] \cdot  \mathbbm{1}[p^u_i[c] < \gamma_{\text{OOD}} ],
\end{equation} 
where $\gamma_{\text{ood}}$ is a threshold set as 0.2, which prevents us from selecting some ID samples that are wrongly classified as OOD.
Considering that each minibatch contains ID samples and not necessarily OOD samples, we store the recyclable OOD samples of each category in a category-wise first-in-first-out memory queue $\mathcal{B}(\cdot)$.

\textbf{Aliasing OOD Matching.}
In training iterations, we first collect the $\mathcal{A}_c$ of the current minibatch and then match each ID feature in it with a random OOD feature in $\mathcal{B}(c)$ as ID-OOD pairs $\{t_i\}$:
\begin{equation}
t_i= ( z_i; o_i ), z_i \in \mathcal{A}_c, o_i \in \mathcal{B}(c). 
\end{equation}
At the end of each iteration, we update each $\mathcal{B}(c)$ by determining whether there are qualified OOD samples (i.e., $n_{i}(c)=1$) in this minibatch.

\subsection{Soft Orthogonality Regularization}
\label{sec:SOR}
    In this section, we introduce our proposed SOR in detail, which includes two parts, as follows:
    \begin{itemize}
        \item We perform a soft orthogonal decomposition (SOD) on ID-OOD pairs to generate pruned ID features.
        \item We design two losses $\mathcal{L}_{odc}^u$ and  $\mathcal{L}_{odc}^l$, which regularize prediction invariance on original ID features and pruned ID features generated by soft orthogonal decomposition. 
    \end{itemize}

\begin{Proposition}
Feature Orthogonal Decomposition (FOD).
\label{Propos}
Any vector $\mathcal{V}$ in the high-dimensional space can be transformed into two mutually orthogonal vectors $\mathcal{V}_a$ and $\mathcal{V}_b$ along a certain basis vector $\mathcal{U}$ direction, formally:
\begin{equation}
\begin{aligned}
&\mathcal{\vec V} = \mathcal{\vec V}_a + \mathcal{\vec V}_b,\\
&\mathcal{\vec V}_a  =  \vec\varepsilon *||\mathcal{\vec V} * \sin < \mathcal{\vec U},\mathcal{\vec V}>||,  \\
& \mathcal{V}_b = \vec \sigma * ||\mathcal{\vec V} * \cos < \mathcal{\vec U},\mathcal{\vec V}>||, \\
&  s.t.\quad  \vec\varepsilon \perp \mathcal{\vec U}, \vec\sigma  \parallel \mathcal{\vec U} , \quad   || \vec\varepsilon|| =|| \vec\sigma || =1,
\end{aligned}
\end{equation}
where $\epsilon$ and $\sigma$ both are unit vectors, and $<\cdot,\cdot>$ denotes the angle between two vectors, $*$ denotes scalar multiplication of vectors.
\end{Proposition} 

\textbf{Soft Orthogonal Decomposition.}
As shown in Fig.~\ref{Fig.sor},
given ID-OOD pairs $t_i =(z_i, o_i)$, SOD applies soft feature orthogonal decomposition on each ID feature $z_i^c$ along with its matching OOD feature $o_i$.
Then we obtain the pruned ID feature $z_{i,r}^c$, which has less similarity with paired OOD features since the OOD semantic component is pruned out of the original ID feature.
According to proposition \ref{Propos}, the process is formulated as follows:

\begin{equation}
\label{zr}
\begin{aligned}
 &\vec z_i = \vec z_{i,a} + \vec z_{i,b}, \\
 &\vec z_{i,r}   = \vec z_i - \alpha \vec z_{i,b},\\
 &s.t.~~ \vec z_{i,a} \perp \vec o_i , \vec z_{i,b}^c \parallel \vec o_i ,\\
 \end{aligned}
\end{equation}
where $\alpha$ (we set $\alpha$ = 0.8) is a hyperparameter to slow down the drastic changes in the feature space caused by FOT, which named soft orthogonal decomposition (SOD).

With the pruned ID feature $ \vec z_{i,r}$ for the anchor ID image $\vec z_{i}$, we obtain its corresponding probability vector $p_{i,r}$ as follows:
\begin{equation}
\label{idfod}
\begin{aligned}
 p_{i,r} &= \mathcal{F} ( \vec z_{i,r} ).
 \end{aligned}
\end{equation}

\textbf{Orthogonality Regularization Loss.}
Moreover, we design orthogonality regularization loss $\mathcal{L}_{odc}^l$ and  $\mathcal{L}_{odc}^u$ to encourage the predictions of our model to be consistent before and after SOD as:
\begin{equation}
    \begin{aligned}
     \mathcal{L}_{odc}^{l} & = \frac{1}{ \sum_{c=0}^K |\mathcal{A}_c^l|}\sum_{c=0}^K \sum_{ z_i^l  \in \mathcal{A}_c^l } KL(p_{i}^l ,p_{i,r}^l  )  \\ & -\frac{1}{|\mathcal{A}_c^l|} \sum_{ z_i^l  \in \mathcal{A}_c^l }\ln(p_{i,r}^l [c]), \\
     \mathcal{L}_{odc}^u  &= \frac{1}{M}\sum_{c=0}^K \sum_{ z_i^u \in \mathcal{A}_c^u  } KL(p_{i}^u ,p_{i,r}^u  ),
      \end{aligned}
\end{equation}
where $\mathcal{L}_{odc}^{l}$ and $\mathcal{L}_{odc}^u$ are orthogonality regularization losses for labeled and unlabeled data, respectively.
For unlabeled data, the $\mathcal{L}_{odc}^u$ is formulated as the KL divergence between $p_{i}^u$ and $p_{i,r}^u$, while for labeled data, we additionally minimize the cross-entropy between $p_{i,r}^l$ and $y_{i}^{l}$ to utilize the label information.

\subsection{Total Loss} 
\label{totalloss}
    In this section, we describe the training  processing and loss functions in detail. As mention above, we use T2T~\cite{huang2021trash} as our plain baseline.
    
    At pre-training stage, our OSP follows baseline, which learns a K-ways predictor with labeled data and a rotation recognizer~\cite{SpyrosGidaris2018UnsupervisedRL} with all unlabeled data to enhance the representation capacity. For the K-ways prediction branch, $ \mathcal{F}$  calculates a K-dimensional class probability vector $p^l_i = \mathcal{F} \circ \mathcal{G}(x^l_i)$. During training, cross entropy is used to regularize the class probability vectors of labeled images:
    \vspace{-10pt}
     \begin{equation}
    \mathcal{L}_{ce} = -\frac{1}{||\mathcal{D}^l||} \sum_{(x_i^l,y_i^l) \in \mathcal{D}^l }  \log p^l_i[y_i^l] ,
        \vspace{-8pt}
    \end{equation}  
     For rotation recognition, we denote four counterparts images $x_j^u$ generated via rotating  by $(k-1)*90^{\circ} $ as $x_{j,k}^u$, then the rotation prediction head $\mathcal{H}(\cdot)$ is responsible for predicting $x_j^u$ with rotation label $k$ with cross entropy loss, 
    \vspace{-10pt}
        \begin{equation}
            \mathcal{L}_{rot} = -\frac{1}{4*||\mathcal{D}^u||} \sum_{(x_i^u) \in \mathcal{D}^u } \sum_{k=1}^{4}   \log q^l_{i,k}[k] ,
         \vspace{-8pt}
        \end{equation}  
To sum up, the total loss of OSP at the pre-training stage is described as follows:
         \vspace{-10pt}
     \begin{equation}
         \mathcal{L}_{pre} = \mathcal{L}_{ce} + \mathcal{L}_{rot} +  \mathcal{L}_{ood}^{l} ,
         \vspace{-8pt}
     \end{equation}
where $\mathcal{L}_{ood}^{l}$ is used to train the OOD detection module $\mathcal{M(\cdot)}$, referring to \cite{huang2021trash}.

At the fine-tuning stage, we apply our proposed orthogonality regularization losses on the baseline, which aims to prune OOD semantic from ID features. 
Referring to~\cite{huang2021trash}, the fine-tuning loss of baseline is described as follows:
    \begin{equation}
     \mathcal{L}_{t2t} = \underbrace{\mathcal{L}_{ce} + \mathcal{L}_{u}}_{\textrm{Classic SSL Loss } }+  \underbrace{\mathcal{L}_{ood}^{l} +\mathcal{L}_{ood}^{u}}_{\textrm{OOD Detection Loss }}+ \mathcal{L}_{rot} .
   \vspace{-8pt}
   \label{eq:t2t}
  \end{equation}
With our proposed orthogonality regularization losses $\mathcal{L}_{odc}^{l}$ and $\mathcal{L}_{odc}^{u}$, the total loss of OSP at the fine-tuning stage is described as follows:
     \begin{equation}
          \mathcal{L}_{ft} =\mathcal{L}_{t2t} + \underbrace{\mathcal{L}_{odc}^{l} + \mathcal{L}_{odc}^{u} }_{\textrm{Our OSR Loss}} 
          \vspace{-8pt}
 \label{eq:osp}
     \end{equation}
where $\mathcal{L}_{ood}^{l}$ and $\mathcal{L}_{ood}^{u}$ are used to train the OOD detection model $\mathcal{D(\cdot)}$\cite{huang2021trash}.

\begin{table*}[]
\centering

\setlength{\tabcolsep}{2.5pt}
\begin{tabular}{c|cc|cc|cc|cc}
\toprule
Method & \multicolumn{2}{c|}{\textbf{MNIST}} & \multicolumn{2}{c|}{\textbf{CIFAR10}} & \multicolumn{2}{c|}{\textbf{CIFAR100}} & \multicolumn{2}{c}{\textbf{TinyImagetNet}} \\ 
  & $\gamma$=0.3 & $\gamma$=0.6  & $\gamma$=0.3  & $\gamma$=0.6 & $\gamma$=0.3& $\gamma$=0.6  & $\gamma$=0.3& $\gamma$=0.6  \\
    \midrule
Supervised        & 93.2 &  93.2   & 76.3 & 76.3 & 58.6 & 58.6 & 36.5 & 36.5   \\ 
  \midrule
       \multicolumn{9}{c}{\textbf{\textit{Classic SSL Methods}}} \\
UDA\ssymbol{2} \cite{xie2019unsupervised}                    & - & - & 90.7 & 88.3 &  67.1 & 64.5  & - & - \\
 Pi-Model\cite{MehdiSajjadi2016RegularizationWS} & 92.4 & 86.6 & 75.7 &74.5 & 59.4 & 57.9 & 36.9 &36.4 \\
PL\cite{DongHyunLee2022PseudoLabelT} & 90.0& 86.0& 75.8&74.6& 60.2&57.5& 36.6& 35.8  \\
VAT\cite{miyato2018virtual}                 &  94.5   &  90.4 & 76.9 & 75.0 &  61.8 &  59.6 &36.7  &36.3  \\
Fixmatch\cite{sohn2020fixmatch}             & - & - & 81.5 & 80.9 & 65.9 & 65.2  & - & -   \\ 
  \midrule
      \multicolumn{9}{c}{\textbf{\textit{Robust SSL Methods}}} \\
DS3L\cite{guo2020safe}                & 96.8 & 94.5 & 78.1 & 76.9 & - & - & - & - \\
UASD\cite{chen2020semi}                    & 96.2 & 94.3 & 77.6 &  76.0 & 61.8 &58.4  &37.1  & 36.9 \\ 
CL\cite{PaolaCascanteBonilla2020CurriculumLR} &  96.9& 95.6 & 83.2& 82.1&63.6&61.5& 37.3& 36.7 \\

Safe-Students\cite{he2022safe}          & 98.3 & 96.5 & 85.7 & 83.8 &68.4  & 68.2 & 37.7 & 37.1  \\
MTC \cite{yu2020multi}              &  93.7&  88.5& 85.5&  81.7&  63.1  &  61.1 &  37.0& 36.6  \\
T2T \cite{huang2021trash}                   &   99.1 & 98.7 & \textbf{91.6} &   \textbf{89.3} &   70.0  & 68.2 & 39.0 &  35.0 \\
  \midrule
\rowcolor{gray!10} Ours  &  \textbf{99.3\scriptsize \textcolor{red}{(+0.2)}} & \textbf{99.4\scriptsize \textcolor{red}{(+0.7)}} & 90.5\scriptsize \textcolor{gray}{(-1.1)} & 88.2\scriptsize \textcolor{gray}{(-1.1)} &  \textbf{72.4\scriptsize \textcolor{red}{(+2.4)}}  &  \textbf{ 70.9\scriptsize \textcolor{red}{(+2.7)}}& \textbf{52.7\scriptsize \textcolor{red}{(+13.7)}} &  \textbf{52.1\scriptsize \textcolor{red}{(+15.0)}}  \\ 
  \bottomrule
\end{tabular}
\caption{\textbf{Intra-dataset:} ID categories classification accuracy (\%) of different methods on the four datasets. In this paper, the \textbf{bold} numbers denote the best results across all approaches. The \textcolor{red}{(+number)} denotes the absolute improvements.}
\label{tab:compare}
\vspace{-8pt}
\end{table*}
\begin{table*}[]
\centering

\setlength{\tabcolsep}{2.5pt}
\begin{tabular}{c|cc|cc|cc|cc}
\toprule
Method & \multicolumn{2}{c|}{\textbf{TIN}} & \multicolumn{2}{c|}{\textbf{LSUN}} & \multicolumn{2}{c|}{\textbf{Gaussian}} & \multicolumn{2}{c}{\textbf{Uniform}} \\ 
  & $N_l$=250 & $N_l$=1000   & $N_l$=250 & $N_l$=1000  & $N_l$=250 & $N_l$=1000  & $N_l$=250 & $N_l$=1000   \\
    \midrule
       \multicolumn{9}{c}{\textbf{\textit{Classic SSL Methods}}} \\

UDA\cite{xie2019unsupervised}                    & 88.8 & 91.8 & 88.5 & 91.1 & 88.9  & 89.2 & 88.7 & 89.7 \\
MixM~\cite{DavidBerthelot2019MixMatchAH}          & 82.4 & 88.0 & 76.3 & 87.0 & 75.8 & 85.7  & 72.9 & 84.5   \\ 
  \midrule
      \multicolumn{9}{c}{\textbf{\textit{Robust SSL Methods}}} \\
DS3L\cite{guo2020safe}                & - & 70.1 & - & 69.7 & - & 62.9 & - & 62.9 \\
UASD\cite{chen2020semi}                    & 83.6 & - & - &  80.9 & - & -  &-  & -\\ 
MTC\cite{yu2020multi}              &  86.4 &  89.9& 86.7& 90.2 &  87.3  &  89.8 & 85.6 & 89.9  \\
OTCT\cite{luo2021consistency}                &-&91.1 & -& 91.3 & - & 92.3 & - & 91.8         \\
T2T \cite{huang2021trash}                   & 91.5 & 93.3 & 91.1 & 94.4 & 90.8  & 93.6 & 90.0 &  94.1\\
  \midrule
\rowcolor{gray!10} Ours  &  \textbf{92.4\scriptsize  \textcolor{red}{(+0.9)}} & \textbf{93.7\scriptsize  \textcolor{red}{(+0.5)}} & \textbf{91.9\scriptsize  \textcolor{red}{(+0.8)}} & \textbf{94.8\scriptsize  \textcolor{red}{(+0.4)}} &  \textbf{91.0\scriptsize  \textcolor{red}{(+0.2)}}  &  \textbf{93.7\scriptsize \textcolor{red}{(+0.1)}}& \textbf{90.8\scriptsize  \textcolor{red}{(+0.8)}} &  \textbf{94.2\scriptsize \textcolor{red}{(+0.1)}}  \\ 
  \bottomrule
\end{tabular}
\caption{\textbf{Inter-dataset:} ID categories classification accuracy (\%) of different methods on CIFAR10 and other four datasets as OOD.}
\label{tab:domain}
\vspace{-10pt}
\end{table*}

\section{Experiments}
\subsection{Experimental Setup}
  \begin{table}
      \centering
      \begin{tabular}{p{45pt}<{\centering}p{22pt}<{\centering}p{22pt}<{\centering}p{50pt}<{\centering}p{35pt}<{\centering}}
    \toprule
    Dataset   & ID classes & OOD classes & labeled samples $N_l$ & OOD samples\\\midrule
    MNIST& 6&4& 6$\times$10& 30,000$\times \gamma$\\
     CIFAR10 & 6 & 4&6$\times$400& 20,000$\times \gamma$ \\
     CIFAR100 & 50 & 50& 50$\times$100 &20,000$\times \gamma$\\ 
     TinyImageNet & 100 & 100 & 100$\times$100 &40,000$\times \gamma$ \\ \bottomrule
      \end{tabular}
  \caption{Intra-dataset OOD setting details.}
      \label{tab:dataset}
\vspace{-8pt}
  \end{table}
\noindent\textbf{Datasets.} Referring to ~\cite{he2022safe}~\cite{huang2021trash}~\cite{guo2020safe}, we evaluate the effectiveness of our OSP on four widely used datasets: MNIST\cite{2011themnist}, CIFAR10\cite{krizhevsky2009learning}, CIFAR100\cite{krizhevsky2009learning} and TinyImageNet\cite{JiaDeng2009ImageNetAL}.

\noindent \textbf{OOD setting.}
In this paper, we use inter-dataset and intra-dataset OOD settings to verify the superiority of OSP.

\noindent \textbf{(a) Intra-dataset OOD Setting:} Following ~\cite{he2022safe}~\cite{huang2021trash}~\cite{guo2020safe}, we select some categories as ID categories and the rest as OOD categories in MNIST~\cite{2011themnist}, CIFAR10~\cite{krizhevsky2009learning}, CIFAR100~\cite{krizhevsky2009learning} and TinyImageNet (a subset of ImageNet~\cite{JiaDeng2009ImageNetAL}). During training, we random sample labeled and unabeled images for ID categories as ID samples and unlabeled images from OOD categories as OOD samples. For MNIST and CIFAR10, we select first six classes as ID categories. For CIFAR100 and TinyImageNet, we select first 50 classes and 100 classes as ID catrgories, respectively. Moreover, we use the mismatch ratio $\gamma \in [0,1] $ to adjust the ratio of OOD samples in the unlabeled data, which modulates class distribution mismatch. For example, when the mismatch ratio $\gamma$ is 0.3, 30\% unlabeled samples come from unseen classes. The details are shown in Tab.~\ref{tab:dataset}. More details about datasets and settings refers to Appendix.

\noindent \textbf{(b) Inter-dataset OOD setting:} Following ~\cite{huang2021trash}, we random sample ID samples from CIFAR-10 and use other dataset to synthesize OOD samples. Specifically, 10,000 unlabeled images are sampled from each of the TIN dataset, the Large-scale Scene Understanding (LSUN) dataset, Guassian noise dataset, and uniform noise dataset, forming into 4 inter-dataset OOD setting.

\noindent\textbf{Metrics.}
Following~\cite{guo2020safe}~\cite{huang2021trash}~\cite{he2022safe}, we choose the mean accuracy (Acc.) to evaluate the classification performance.
For OOD detection, we use the area under the receiver operating characteristic (AUROC) as metrics\cite{huang2021trash}.

\noindent\textbf{Implementation Details.}
Existing methods including UDA \cite{xie2019unsupervised}, FixMatch \cite{sohn2020fixmatch}, VAT \cite{miyato2018virtual}, PL \cite{DongHyunLee2022PseudoLabelT}, Pi-Model \cite{MehdiSajjadi2016RegularizationWS}, MTC \cite{yu2020multi}, DS3L \cite{guo2020safe}, UASD \cite{chen2020semi}, CL \cite{PaolaCascanteBonilla2020CurriculumLR}, T2T \cite{huang2021trash}  and Safe-Student \cite{he2022safe} are used for comparison. For our method, SGD is used to optimize network weights. The learning rate is initially set to 0.03 at the pre-training stage and  0.001  at the fine-tuning stage, which is adjusted via the cosine decay strategy\cite{xie2019unsupervised,sohn2020fixmatch}. The momentum is set to 0.9. In each training batch, the batch size of labeled data and unlabeled data are 64 and 320. And the pre-training stage costs 50,000 iterations, and the fine-tuning stage takes 200,000 iterations. We set the size of recyclable OOD Bank $\mathcal{B}(\cdot)$ is 5000. For  UDA\cite{xie2019unsupervised} and FixMatch \cite{sohn2020fixmatch}, models are trained with 250,000 iterations for a fair comparison.For far comparison, when training MTC\cite{yu2020multi}  and T2T \cite{huang2021trash}, we follow their original settings in \cite{yu2020multi} and  \cite{huang2021trash}, respectively. In MNIST, we adopt a simple two-layer CNN model as a backbone network\cite{he2022safe}\cite{guo2020safe}, while in CIFAR10, CIFAR100 and TinyImageNet, we use the Wide-ResNet28-2\cite{SergeyZagoruyko2016WideRN} as the backbone model. 

\begin{table}[t]
    \centering
    \begin{tabular}{ccc|c|c}
 \toprule
 $ \mathcal{L}_{odc}^{u}$  & $ \mathcal{L}_{odc}^{l}$ &  AOM & \textbf{TinyImageNet}  & \textbf{CIFAR100}  \\ \midrule

  &     &    & 35.0 & 68.2  \\
  & \CheckmarkBold    &  \CheckmarkBold  &  49.5& 69.9  \\%
   
  \CheckmarkBold     &    &  \CheckmarkBold   &  48.4   & 70.4  \\
  \CheckmarkBold   &  \CheckmarkBold    &        &  46.5 & 70.5 \\
 \rowcolor{gray!10}      \CheckmarkBold   & \CheckmarkBold &  \CheckmarkBold   &  \textbf{52.1} & \textbf{69.1} \\
\bottomrule
    \end{tabular}
    \caption{Abaltion results on CIFAR100 ($\gamma=0.6$) and TinyImageNet ($\gamma=0.6$) }
    \label{tab:ablation}
\vspace{-10pt}
\end{table}
\begin{table}[t]
\centering
\centering
\setlength{\tabcolsep}{4 pt}
\begin{tabular}{ccccc}
 \toprule 
 Method & MNIST & CIFAR10 & CIFAR100 &  TinyImageNet   \\ \midrule
   T2T~\cite{huang2021trash} & 92.6 & 67.4 & 64.8 & 40.5  \\
  \rowcolor{gray!10} Ours & \textbf{99.8}  & \textbf{88.3} & \textbf{71.8} & \textbf{54.4}  \\
 \bottomrule
\end{tabular} %
\caption{The OOD detection performance comparison accross different datasets ( AUROC(\textit{\%})).
}
\label{tab:ood}
 \vspace{-20pt}
\end{table}

\begin{figure*} 
\centering 
  \vspace{0.1pt}
\includegraphics[width=0.8\textwidth]{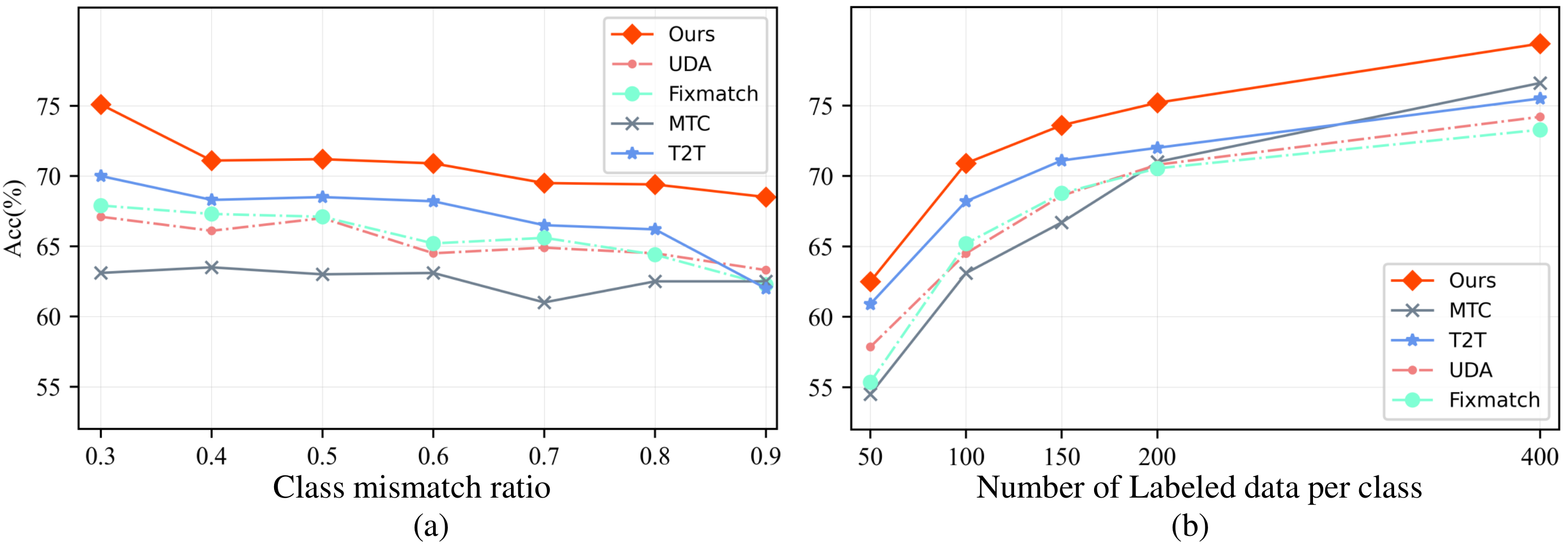} 
\vspace{-8pt}
\caption{(a) Effect of the class mismatch ratio. (b) Effect of the labeled data amount.
All these results are obtained on the CIFAR100 dataset with 100 labeled data per class.
} 
  \label{Fig.ratio}
\vspace{-15pt}
\end{figure*}

\subsection{Main Results.}
\textbf{OOD proportion of datasets.} 
Here, we report the proportion of OOD samples in different datasets to help understand the performances of OSP.
As Tab.~\ref{tab:dataset} shows, hard datasets like TinyImageNet contain more OOD classes and samples, for which obtaining a clear ID/OOD discrimination is very hard.
In other words, the `feature aliasing' problem corrupts learning more heavily on hard datasets (e.g., TinyImageNet) than on easy ones (e.g., CIFAR10).

\textbf{Performance on intra-dataset setting.}
As shown in Tab.~\ref{tab:compare}, our OSP achieves the best performance on MNIST, CIFAR100, and TinyImageNet with various class mismatch ratios $\gamma$.
Prominently, on TinyImageNet, most existing methods have low accuracy but our OSP improves the best baseline by \textbf{13.7\%} and \textbf{15.0\%} when the class mismatch ratio $\gamma=0.3$ and $0.6$, respectively.
This is because our OSP is designed to tackle the "feature aliasing” problem, and this problem matters heavily in hard datasets like TinyImageNet as mentioned above.
While for easy datasets, our OSP also obtains competitive performances to SOTA alternatives.
These comparisons highlight the superiority of OSP in addressing the corruption from OOD data.

\textbf{Performance on inter-dataset setting.}
As shown in Tab.~\ref{tab:domain}, OSP outperforms previous methods on CIFAR10 with various OOD datasets (e.g. TIN, LSUN, Gaussian, and Uniform).
This indicates the good versatility of OSP for different OOD sources, reflecting its potential in real complex dataset settings.

\textbf{Results on various class mismatch ratio.}
To verify the robustness of our OSP to corruption of unlabeled data, we illustrate the performance of our model under various mismatch ratios in CIFAR100 with 100 labeled data per class. The results are shown in Fig.~\ref{Fig.ratio}(a). We see that our OSP achieves SOTA in all settings. Moreover, most baselines display significant performance degradation as $\gamma$ increases, whereas OSP remains competitive. These observations clearly validate the superiority of OSP.

\textbf{Results on different labeled data amount.} Moreover, we further verify the effectiveness of our OSP under different labeled data amounts. Here, we carry out all experiments on CIFAR100 with $\gamma=0.6$. As shown in Fig.~\ref{Fig.ratio}(b), our OSP obtains the best performances on all labeled data amount settings, reflecting the broad applicability of our approach.
A notable point is that the advantages of previous robust SSL methods~\cite{huang2021trash}~\cite{yu2020multi} gradually fade away with the increase in the amount of labeled data.

\begin{figure*}[t] 
\centering 
\includegraphics[width=0.7\textwidth]{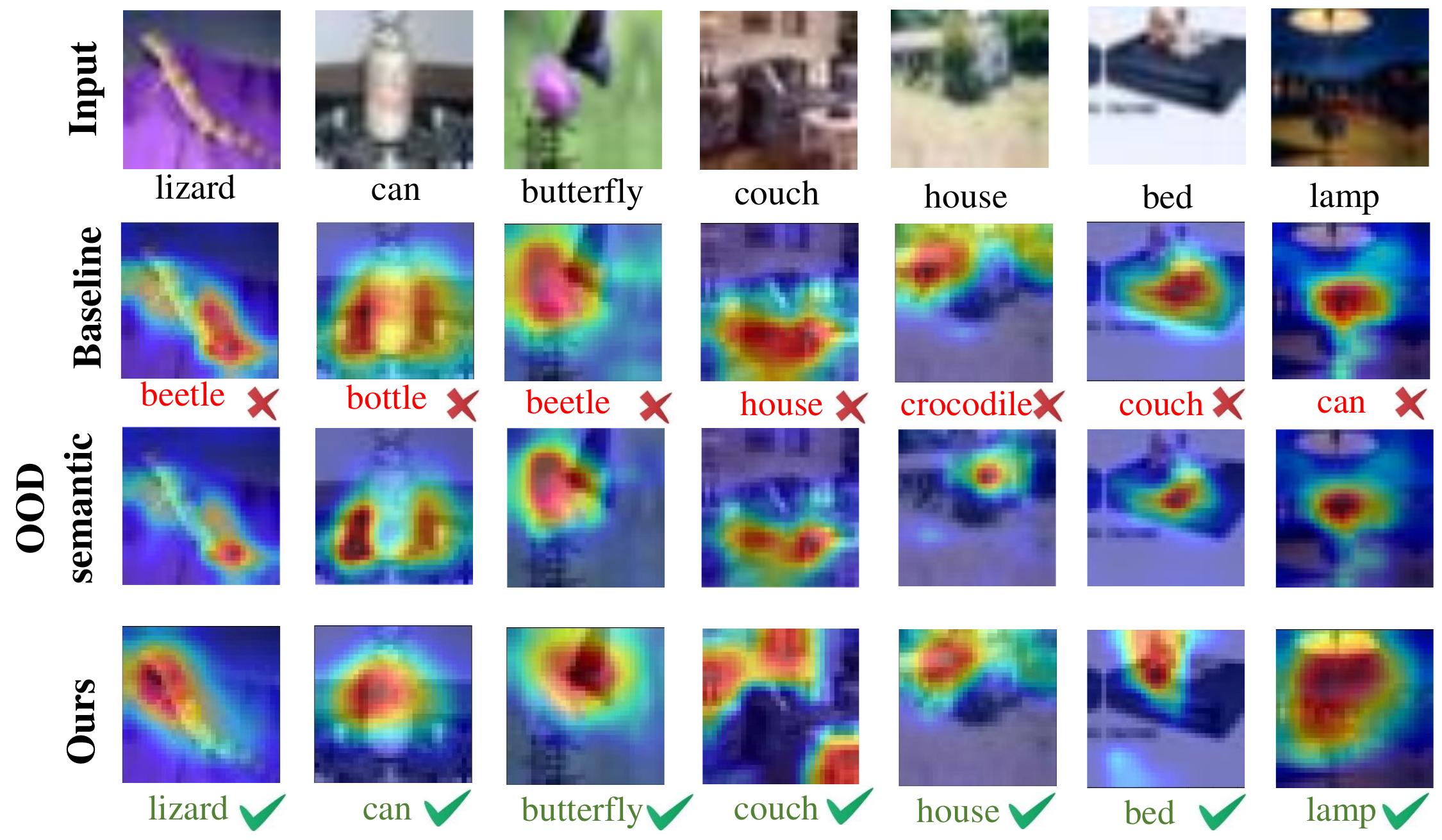} 
\caption{Activation maps of baseline~\cite{huang2021trash} and OSP using Grad-CAM\cite{RamprasaathRSelvaraju2016GradCAMVE}.
The red (blue) color represents more (less) attention from the model.
Rows 1-4 represent input images, CAMs from baseline\cite{huang2021trash}, paired OOD features in OSP, and CAMs from OSP, respectively.
} 
\label{Fig.cam} 
\end{figure*}
\subsection{Ablation Studies}
\textbf{Effect of Soft Orthogonality Regularization.} To verify the effectiveness of our SOR, we compare four variants: (1) Row 1: the baseline without our proposed AOM and SOR and use Eq.\ref{eq:t2t} as finetuning loss function. 
(2) Row 2: only applies SOR on labeled ID anchor features $\mathcal{A}_c^l$.  
(3) Row 3: only applies SOR on unlabeled ID anchor features $\mathcal{A}_c^u$. (4) Row 5: our OSP which applies SOR on all ID anchor features $\mathcal{A}_c^u \cup \mathcal{A}_c^l $.  
As shown in Tab.~\ref{tab:ablation}, our SOR module outperforms baseline obviously and our proposed regularization loss $\mathcal{A}_c^l$ and $\mathcal{A}_c^u$ both contribute to performance improvements.

\textbf{Effect of Aliasing OOD Matching.} To quantify the impact of AOM, we compare two variants: (1) Row 4: random selects OOD features to pair ID features (2) Row 5: our OSP which matches each ID sample with an OOD sample that has a large semantic overlap with it, as described in Sec.~\ref{sec:ASMM}. From Tab.~\ref{tab:ablation}, the results indicate that our ID-OOD pairs procedure (AOM) is beneficial to pruning OOD semantic and further improves performance.

\subsection{Further Analysis}

\textbf{OOD detection.}
    In Tab.~\ref{tab:ood}, we compare our method against T2T~\cite{huang2021trash} under combinations of ID and OOD datasets, to validate the efficacy of our OSP . The AUROC is used as the metric here.
    We see that our OSP outperforms T2T~\cite{huang2021trash} under all settings with a large margin. reflecting the superority of OSP in ID/OOD discrimination.

\textbf{Visualization of class activation map.}
We use Grad-CAM \cite{RamprasaathRSelvaraju2016GradCAMVE} to visualize the class activation map. As shown in Fig.~\ref{Fig.cam}, we notice that the baseline (row 2) is distracted and even focuses on non-foreground object regions, thus has wrong predictions.
In contrast, OSP focuses on the object regions more accurately and comprehensively (row4), indicating the superiority of OSP in learning semantic structure.
This is because OSP encourages our model to only reserve classification-related ID semantics by pruning classification-useless OOD semantics, which is mostly activated in the background region (row 3).

\textbf{More results on real-world dataset.}
    STL-10~\cite{coates2011analysis} is a dataset for real-world image recognition, while each class has fewer labeled training examples (ID samples) and a very large set of unlabeled OOD examples. The unlabeled OOD samples comes from a similar but different distribution from the labeled data. The primary challenge is to make use of the unlabeled data to improve recognition for the ID samples. Here, we resize the images as 32$\times$32. We applied our OSP on STL-10 with 20,000 OOD samples, 100 labeled and 200 unlabeled ID samples per class. Our OPS improves T2T~\cite{huang2021trash} by 3.1\% (Acc.78.0\% v.s. 74.9\%).

\section{Conclusion}
    In this paper, we introduce a novel method named OSP for robust semi-supervised learning~\cite{zhao2020robust,LanZheGuo2022RobustDS}, which first exploits the value of OOD at the semantic level.
    Our OSP mitigates the corruption from OOD samples by pruning OOD semantics out from ID features at the semantics level.
    Specifically, we propose an aliasing OOD matching module to pair each ID sample with an OOD sample with which it has semantic overlap.
    We then develop a soft orthogonality regularization to regularize the ID and OOD features to be orthogonal.
    Further, we will extend our OSP to more challenging open-set scenarios~\cite{DanHendrycks2016ABF,KiminLee2018ASU,JaehyungKim2020DistributionAR}.
    
\noindent \textbf{Acknowledgements.}
This work was supported in part by the National Key R\&D Program of China (No.2022ZD0118201), Natural Science Foundation of China (No.61972217, 32071459, 62176249, 62006133, 62271465), and the Natural Science Foundation of Guangdong Province in China (No.2019B1515120049).

{\small
\bibliographystyle{ieee_fullname}
\bibliography{main}
}

\newpage
\title{Out-of-Distributed Semantic Pruning for Robust Semi-Supervised Learning Supplementary Materials}


\maketitle
\appendix

\begin{figure*}[h] 
  \centering
    \includegraphics[width=0.95\textwidth]{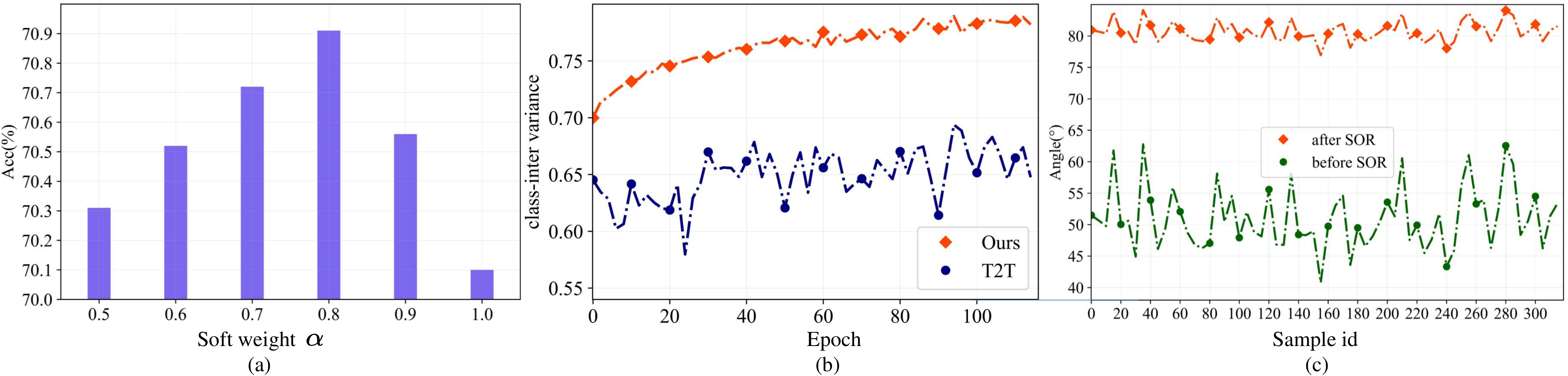} 
\caption{(a) The ablation study about soft weight $\alpha$. (b) The class-inter variance between ID features. (c) The angle between ID and OOD features. All these results are obtained on CIFAR100 with 100 labeled data per class and $\gamma$=0.6.}
  \label{Fig.variance}
\end{figure*}

\section{The performance of OOD detection}
    To further measure the potential to identify OOD classes, we compare our OSP with T2T~\cite{huang2021trash} in CIFAR100 with 100 labeled data per class. We utilize AUROC \cite{huang2021trash,he2022safe} to evaluate performance for OOD detection. The results are shown in Table.~\ref{stb:ablation}. Our OSP  surpasses the baseline~\cite{huang2021trash} under all setting. This indicates that our OSP promotes OOD detection by keeping ID features and OOD semantic orthogonal. 
\begin{table}[h]
\centering
\setlength{\tabcolsep}{5pt}
\begin{tabular}[]{cccccccc}
 \toprule
& \multicolumn{7}{c}{Class Mismatch Ratio} \\
 Method &0.3  & 0.4 &  0.5 & 0.6  & 0.7 &0.8 & 0.9 \\ \midrule
T2T~\cite{huang2021trash} & 65.5 & 61.0&59.0 &60.9 &56.3 &56.7 &59.4 \\
\rowcolor{gray!10}  Ours & 68.3& 73.2&69.4 &69.1 &66.7 &64.3 & 63.0
\\ \bottomrule
\end{tabular} %
\caption{AUROC(\%) for OOD detection on CIFAR100 with 100 labeled data per class.}
\label{stb:ablation}
\end{table}

\section{Ablation Study}

\textbf{Effect of soft weight $\bm{\alpha}$}.
   The parameter $\alpha$ is a hyper-parameters to adjust the drastic changes in the feature space caused by the orthogonal operation (in Eq.~\ref{zr}).
   We conduct ablation experiments on different soft weights $\alpha$ to explore the effect of it on OSP.
   As shown in Fig.~\ref{Fig.variance}(a), small $\alpha$ weakens the effect of our OSP, while large $\alpha$ leads to dramatic changes in feature space.
   Given our observation of the trade-off, we adopt $\alpha = 0.8$ in all our experiments.

\section{Further Analysis}
\textbf{Analysis of the angle of ID and OOD features.}
As shown in Fig.~\ref{Fig.variance} (b), we see the baseline T2T has an angle around $50^{\circ}$, the cosine similarity is $64\%$ (i.e., $\cos(50^{\circ})$), which means there is an amount of meaning aliasing between ID and OOD features.
In contrast, the feature angles after our OSP are around $80^{\circ}$, which remarkably suppresses their similarity to about only $17\%$ (i.e., $\cos(80^{\circ})$). This indicates that our model effectively prunes OOD semantics out from ID features, enhancing the discrimination of ID and OOD samples.

\textbf{Analysis on the inter-class variance.}
As shown in Fig.~\ref{Fig.variance}~(c), our OSP obtains a significantly larger inter-class variance than the baseline\cite{huang2021trash}, reflecting OSP obtains inter-class discrimination with higher generalizability~\cite{ShaofengZhang2022AlignRW}.
Another interesting property is that OSP encourages the inter-class variance to increase within training, whereas the baseline\cite{huang2021trash} does not. This suggests that OSP progressively acquires discriminative ID class semantics during training.

\section{Datasets}

\subsection{intra-dataset setting}
    For intra-dataset setting, we follow~\cite{he2022safe}~\cite{guo2020safe} to evaluate OSP on image classification datasets: MNIST~\cite{2011themnist}, CIFAR10~\cite{krizhevsky2009learning}, CIFAR100~\cite{krizhevsky2009learning} and TinyImageNet(a subset of ImageNet~\cite{JiaDeng2009ImageNetAL}), with different class mismatch ratio $\gamma$.
    
    \textbf{MNIST} includes 60,000 training images and 10,000 testing images of size 28 × 28, which contains 10 categories from digit \textit{0} to digit \textit{9}. In this paper, we consider the first six categories (from digit \textit{0} to digit \textit{5}) as $\mathcal{C}^l$ and the remaining categories as OOD categories, $\mathcal{C}^{ood}$. Moreover, we respectively select ten images from $\mathcal{C}^l$ to construct the labeled data set $\mathcal{D}^l$, i.e., a total of 60 labeled data, and select 30,000 images in total from digit \textit{0} to digit \textit{9} as unlabeled data $\mathcal{D}^u$ .Moreover, we use the mismatch ratio $\gamma$ to adjust the ratio of OOD samples in the unlabeled data to modulate class distribution mismatch. For example, when the extent of labeled/unlabeled class mismatch ratio is 0\%, all unlabeled data come from digit \textit{0}  to digit \textit{5}.
    
    \textbf{CIFAR10} includes 60,000 training images and 10,000 testing images of size 32 × 32 which contains ten categories: \textit{airline, automobile, bird, cat, deer, dog, frog, horse, ship} and \textit{trunk}. Our experiment carries out six-categories classification tasks. We consider animal categories (\textit{birds, cats, deer, dogs, frogs and horses}) as ID categories and the rest as OOD categories. We select 400 images from each ID category to construct the labeled data set $\mathcal{D}^l$, i.e., 2400 labeled instances. Meanwhile, 20,000 images in total are randomly selected as the unlabeled data set $\mathcal{D}^u$ from all the ten categories. We adjust the ratio of OOD images in the unlabeled data to modulate class distribution mismatch $\gamma$.

    \textbf{CIFAR100} includes 50,000 training images and 10,000 testing images of size 32 × 32 which contains 100 categories. We use the first half categories (1-50) as ID categories, and the remaining classes as OOD categories. We select 100 images from each ID category to construct the labeled data set  $\mathcal{D}^l$ , i.e., 5000 labeled instances. Meanwhile, 20,000 images in total are randomly selected as the unlabeled data set  $\mathcal{D}^u$ from all the 100 categories with different ratios of OOD classes.

    \textbf{TinyImagetNet} contains 200 categories which includes 500 training images and 50 testing images in each category. We resize all images to 32 × 32. We use the first 100 categories as ID classes, and the remaining classes as OOD categories. We select 100 images from each ID category to construct the labeled data set $\mathcal{D}^L$ , i.e., 10000 labeled instances. Meanwhile, 40,000 images in total are randomly selected as the unlabeled data set $\mathcal{D}^u$  from all the 200 categories with different ratios of OOD classes.

\subsection{inter-dataset setting}
    For inter-dataset setting, we follow~\cite{huang2021trash} to evaluate OSP on CIFAR10 with different amounts of labeled data. Here, CIFAR10 is used as ID samples, and we samples 10,000 images from other dataset as OOD samples, e.g. the TIN dataset, the Large-scale Scene Understanding (LSUN) dataset~\cite{yu2015lsun}, Gaussian noise dataset, and uniform noise dataset, forming into our inter-dataset settings. For CIFAR10, following the original split, 10,000 images are used for testing and the same splits in~\cite{huang2021trash}~\cite{yu2020multi} are adopted for training and validating.

\section{Algorithm}
    We provide our training algorithm in Alg.~\ref{alg: alogirthm1}. The training processing consists of two stages: pre-training stage and fine-tuning stage.

\begin{algorithm*}[h]
\caption{Training algorithm}
\label{alg: alogirthm1}
\textbf{Input:} Labeled data, $\mathbb{D}^l$, a set of unlabeled data $\mathbb{D}^u$, an encoder $\mathcal{G}(\cdot)$, a K-ways classifier $\mathcal{F}(\cdot)$, a rotation prediction head $\mathcal{H}(\cdot)$, an OOD detection module $\mathcal{M}(\cdot)$, pre-training epochs $E_1$, Fine-tuning epochs $E_2$, max iteration per epochs $I$, temperature $T$.  \\
\textbf{Output:} Trained encoder $\mathcal{G}(\cdot)$ and Trained K-ways classifier $\mathcal{F}(\cdot)$.
\\
\begin{algorithmic}[1] \\
\textbf{/*********************** Pre-training Stage ************************/}
\For{{$e$} = 1...{$E_1$}}  
  \For{{$i$} = 1...{$I$}}\\
 $\quad\quad$  \textbf{compute} $ \mathcal{L}_{pre} = \mathcal{L}_{ce} + \mathcal{L}_{rot}+ \mathcal{L}_{ood}^l$ \Comment{Eq.~13} \\ 
 $\quad\quad$  \textbf{update}  $\mathcal{G}(\cdot)$,$\mathcal{F}(\cdot)$, $\mathcal{H}(\cdot)$,$\mathcal{M}(\cdot)$  $\longleftarrow $ SGD with $ \mathcal{L}_{pre}$.
 \EndFor
\EndFor
\\
\textbf{/*********************** Fine-tuning Stage ************************/} \\
\textbf{Initialize OOD samples set  $\mathbb{U}^{ood} =\emptyset $ }.
\For{{$e$} = 1...{$E_2$}}
    \If{$e\%10=0$}  \\ 
 $\quad\quad$    \textbf{ update }  $\mathbb{U}^u$ , $\mathbb{U}^{ood}$ $\longleftarrow$ \textbf{split old unlabeled data }  $\mathbb{D}^u$ \textbf{ with } $\mathcal{M}(\cdot)$. \Comment{Eq.~6}
 \EndIf
  \If{$\mathbb{U}^{ood} \ne \emptyset $}  \\
 $\quad\quad$ \textbf{ select recyclable OOD samples } for $\mathbb{U}^{ood}$  \textbf{ and update recyclable OOD Bank } $\mathcal{B}(c)$ ($c=1...K$).
 \Comment{Eq.~5} \\
  \EndIf
  \For{{$i$} = 1...{$I$}}
 \State $(\mathbf{B}_l=\{x^l,y^l\}, \mathbf{B}_u=\{x^u\})  \gets \mathrm{SampleBatch}(\mathbb{D}^l, \mathbb{D}^u)$.\\
 $\quad\quad\quad$  $ {z_i^l = \mathcal{G}(x_i^l)}$ and  $ {z_j^u = \mathcal{G}(x^u_j)}$,where, $x_i^l \in \mathbf{B}_l,x^u_j \in \mathbf{B}_u $.
 \For{{$c$} = 1...{$K$}}     $\quad\quad$ // compute anchor ID samples set \\
    $\quad\quad\quad\quad$ $\mathcal{A}^l_c = \{z^l_i | z^l_i = \mathcal{G}(x^l_i),y^l_i = c, p^l_i[c] > \delta \}$, \Comment{Eq.~2} \\
    $\quad\quad\quad\quad$ $\mathcal{A}^u_c = \{z^u_i | z^u_i = \mathcal{G}(x^u_j),\hat y^u_j = c, p(z^u_j) > \delta \}$, \Comment{Eq.~3} \\
    $\quad\quad\quad\quad$   $\mathcal{A}_{c} = \mathcal{A}^l_c \cup \mathcal{A}^u_c $.  \Comment{Eq.~4}
 \EndFor
  \textbf{obtain all anchor ID samples set} $ \mathcal{A} = \{ \mathcal{A}_{c} \}_{c=1}^K$. \\
 
 $\quad\quad\quad$\textbf{compute ID-OOD pairs} $t_i$ for each $z_i$ in $\mathcal{A}$  $\quad\quad$ // our AOM module \Comment{Eq.~6} \\ 
 
$\quad\quad\quad$\textbf{compute pruned ID features} $z_{i,r}$ for each $t_i$ in $\mathcal{A}$. $\quad\quad$ // our SOT module \Comment{Eq.~8,9}  \\ 

 $\quad\quad\quad$\textbf{compute} $\mathcal{L}_{ft} = \underbrace{\mathcal{L}_{ce} + \mathcal{L}_{u}}_{\textrm{Classic SSL Loss } }+  \underbrace{\mathcal{L}_{ood}^{l} +\mathcal{L}_{ood}^{u}}_{\textrm{OOD Detection Loss }}+ \underbrace{\mathcal{L}_{odc}^{l} + \mathcal{L}_{odc}^{u} }_{\textrm{Our OSR Loss}}+ \mathcal{L}_{rot}$\Comment{Eq.~15} \\ \\
 $\quad\quad\quad$\textbf{update}  $\mathcal{G}(\cdot)$,$\mathcal{F}(\cdot)$, $\mathcal{H}(\cdot)$,$\mathcal{M}(\cdot)$  $\longleftarrow $ SGD with $ \mathcal{L}_{ft}$.
    \EndFor
\EndFor
\State \textbf{return} encoder $\mathcal{G}(\cdot)$ and K-ways classifier $\mathcal{F}(\cdot)$

\end{algorithmic}
\end{algorithm*}

\end{document}